\documentclass[letterpaper]{article} 
\usepackage{aaai2026}
\usepackage{times}  
\usepackage{helvet}  
\usepackage{courier}  
\usepackage[hyphens]{url}  
\usepackage{graphicx} 
\usepackage{natbib}  
\usepackage{caption} 
\usepackage{algorithm}
\usepackage{algorithmic}
\usepackage{amsmath}
\usepackage{amssymb}
\usepackage{booktabs} 
\usepackage{multirow}
\usepackage{graphicx}
\usepackage{subcaption}
\usepackage{float}
\usepackage{xcolor}
\usepackage{newfloat}
\usepackage{listings}
\DeclareCaptionStyle{ruled}{labelfont=normalfont,labelsep=colon,strut=off} 
\floatstyle{ruled}
\newfloat{listing}{tb}{lst}{}
\floatname{listing}{Listing}
\nocopyright 

\title{CIVQLLIE: Causal Intervention with Vector Quantization for Low-Light Image Enhancement}
\author{
    Tongshun Zhang\textsuperscript{\rm 1,\rm 2},
    Pingping Liu\textsuperscript{\rm 1,\rm 2}\thanks{Corresponding author},
    Zhe Zhang\textsuperscript{\rm 1,\rm 2},
    Qiuzhan Zhou\textsuperscript{\rm 3}
}
\affiliations{
    \textsuperscript{\rm 1} College of Computer Science and Technology, Jilin University\\
    \textsuperscript{\rm 2} Key Laboratory of Symbolic Computation and Knowledge Engineering of Ministry of Education, Jilin University\\
    \textsuperscript{\rm 3} College of Communication Engineering, Jilin University\\

    \{tszhang23, zhezhang23\}@mails.jlu.edu.cn,\{liupp, zhouqz\}@jlu.edu.cn
}

\usepackage{bibentry}

\begin{document}

\maketitle

\begin{abstract}
Images captured in nighttime scenes suffer from severely reduced visibility, hindering effective content perception. Current low-light image enhancement (LLIE) methods face significant challenges: data-driven end-to-end mapping networks lack interpretability or rely on unreliable prior guidance, struggling under extremely dark conditions, while physics-based methods depend on simplified assumptions that often fail in complex real-world scenarios. To address these limitations, we propose CIVQLLIE (\textbf{C}ausal \textbf{I}ntervention on \textbf{V}ector \textbf{Q}uantization for Low-Light Image Enhancement), a novel framework that leverages the power of discrete representation learning through causal reasoning. We achieve this through Vector Quantization (VQ), which maps continuous image features to a discrete codebook of visual tokens learned from large-scale high-quality images. This codebook serves as a reliable prior, encoding standardized brightness and color patterns that are independent of degradation. However, direct application of VQ to low-light images fails due to distribution shifts between degraded inputs and the learned codebook. Therefore, we propose a multi-level causal intervention approach to systematically correct these shifts. First, during encoding, our Pixel-level Causal Intervention (PCI) module intervenes to align low-level features with the brightness and color distributions expected by the codebook. Second, a Feature-aware Causal Intervention (FCI) mechanism with Low-frequency Selective Attention Gating (LSAG) identifies and enhances channels most affected by illumination degradation, facilitating accurate codebook token matching while enhancing the encoder's generalization performance through flexible feature-level intervention. Finally, during decoding, the High-frequency Detail Reconstruction Module (HDRM) leverages structural information preserved in the matched codebook representations to reconstruct fine details using deformable convolution techniques. Extensive experiments demonstrate that our approach achieves superior recovery performance on both standard benchmarks and real-world low-light images. Code is available at 
https://github.com/bywlzts/CIVQLLIE. 
\end{abstract}

\begin{figure*}[t]
    \centering
    \includegraphics[width=1\linewidth]{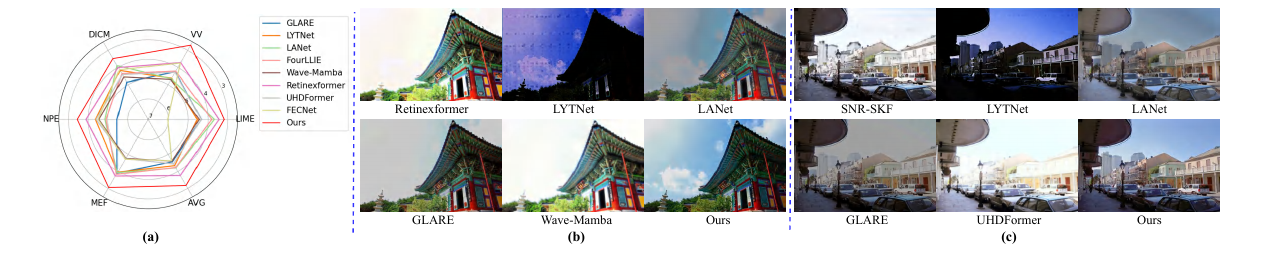}
    \vspace{-0.7cm}
    \caption{Qualitative and Quantitative Comparisons with Recent SOTA Methods. (a) illustrates the NIQE comparison results on five unpaired natural datasets, where AVG represents the average performance. (b) and (c) present visual comparisons with recent state-of-the-art methods. These results validate the robust generalization capability of our method in real-world scenarios.}
    \vspace{-0.2cm}
    \label{fig:page1}
\end{figure*}

\section{Introduction}
LLIE enhances visibility in dark scenes, recovering vital information for computer vision tasks such as object detection~\cite{hashmi2023featenhancer, du2024boosting}, scene understanding~\cite{laskar2025dataset}, and autonomous driving~\cite{li2024light}. Early LLIE methods, including histogram equalization~\cite{pisano1998contrast} and Retinex theory~\cite{ng2011total}, relied on image statistics and physical models. However, these techniques often fall short in complex lighting situations, leading to unnatural color shifts and detail distortions in low-light or intricate scenes.

Recent advancements in deep learning have accelerated LLIE research. End-to-end CNN approaches~\cite{cai2023retinexformer, wang2023fourllie, feijoo2025darkir, yan2025hvi, zhang2025cwnet} learn direct mappings from low-light inputs to enhanced outputs but face three inherent limitations: (1) black-box nature limits interpretability, (2) poor generalization under extreme low-light conditions, and (3) artifacts in color reproduction and detail recovery with unseen illumination patterns.
Alternative methods using physical priors~\cite{wei1808deep, cai2023retinexformer, bai2024retinexmamba} or auxiliary guidance~\cite{zhang2024dmfourllie, wu2023learning} attempt to address these issues but are often constrained by oversimplified degradation assumptions or unreliable prior estimations derived from degraded inputs.

To address these limitations, we employ vector quantization (VQ) codebooks as discrete repositories of standardized brightness and color patterns learned from high-quality normal-light images. Unlike traditional approaches relying on hand-designed features or unreliable degradation priors, our VQGAN-based framework~\cite{esser2021taming} captures intrinsic visual patterns through data-driven quantization of large-scale normal-light imagery. 
VQGAN-based methods have demonstrated their effectiveness in various low-level vision tasks, including LLIE~\cite{liu2023low,zou2024vqcnir,wang2024perceplie}, super-resolution~\cite{chen2022real}, and dehazing~\cite{wu2023ridcp,fu2025iterative}. However, when directly applying VQ codebooks to process extremely low-light images, a significant feature domain mismatch occurs—insufficient illumination causes low-light image features to deviate substantially from normal-light features stored in the codebook, leading to noise amplification and artifact generation that affect reconstruction quality. Therefore, establishing precise mapping relationships between low-light degraded features and high-quality codebook features becomes the key challenge for achieving efficient dark scene enhancement.

Unlike existing codebook-based image restoration methods~\cite{liu2023low,zou2024vqcnir,wang2024perceplie}, this paper re-examines LLIE through the lens of causal analysis, proposing CIVQLLIE. This framework thoroughly analyzes dark scene image characteristics and implements carefully designed causal interventions during the VQGAN encoder-codebook pipeline. First, we establish and adhere to causal intervention principles for LLIE, designing a Pixel-level Causal Intervention (PCI) module that operates in the pixel domain, selectively intervening on illumination and chrominance components while preserving structural information. This causal intervention effectively guides degraded features toward normal illumination and color distributions encoded in the VQ codebook. Furthermore, to enable more flexible causal intervention in high-dimensional encoded features, we propose a Feature-level Causal Intervention (FCI) mechanism that incorporates Low-frequency Selective Attention Gating (LSAG) to precisely identify channels highly correlated with brightness and color. 
This enables interventions to occur in channels where brightness and color are highly correlated. Meanwhile, feature-level interventions enhance the adaptability to various low-light scenarios in the real world. 
In the decoding stage, we introduce a High-frequency Detail Refinement Module (HDRM) that utilizes high-frequency sensitive channels (complementary to low-frequency sensitive channels) as guidance to precisely adjust encoder features through deformable convolution techniques, achieving high-quality detail and structure reconstruction. 

Fig.\ref{fig:page1} provides both qualitative and quantitative evidence showcasing the SOTA performance of our approach on real-world low-light images. Our method exhibits robust generalizability, excellent visual quality, and effective image detail recovery. Overall, the main contributions of this paper can be summarized as follows:   
\vspace{-0.1cm}
\begin{itemize}  
    \item We introduce a novel causal analysis framework that bridges the gap between degraded features and high-quality representations in vector quantization codebooks, addressing key limitations in VQGAN-based restoration.
    \item We propose a dual-stage intervention strategy comprising PCI and FCI within a dedicated causal framework.  
    \item We develop the LSAG to target illumination-sensitive channels and with a HDRM that integrates fine details through deformable convolutions, enhancing both global illumination correction and local structure preservation.  
    \item Our extensive experiments show that CIVQLLIE achieves state-of-the-art performance on standard LLIE benchmarks and demonstrates superior generalization on challenging real-world low-light scenes.
\end{itemize}

\section{Method}
\subsection{Preliminaries}

\begin{figure}[t]
    \centering
    \includegraphics[width=0.9\linewidth]{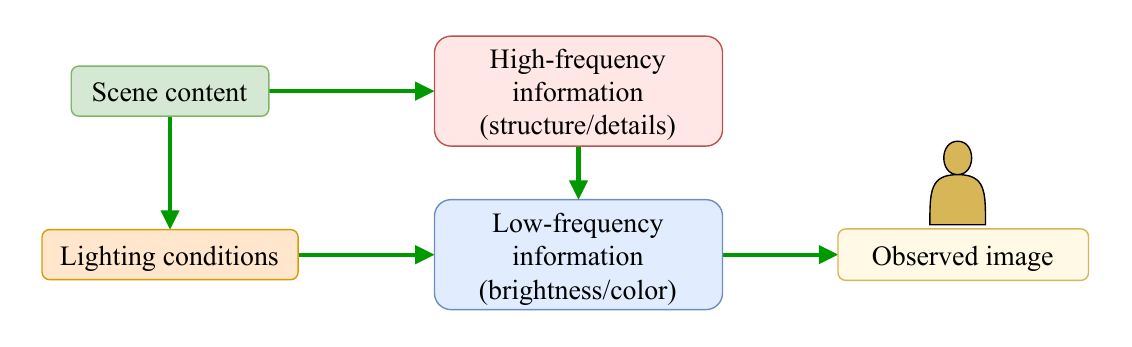}
    \vspace{-0.3cm}
    \caption{Visualization of the causal DAG for LLIE.}
    \vspace{-0.3cm}
    \label{fig:dag}
\end{figure}

\textbf{LLIE Causal Intervention Principles.} We propose a novel perspective by integrating causal inference principles with metric learning to construct a latent space that precisely captures the true impact of lighting conditions. From a causal standpoint, image quality degradation in LLIE can be attributed to specific causal chains, as illustrated in Fig.\ref{fig:dag}, where we construct a directed acyclic graph (DAG) to represent this process. From human visual perception, the difficulty in recognizing low-light images primarily stems from insufficient illumination, which directly affects the low-frequency features of the image. Therefore, we define illumination-affected low-frequency degradation factors (brightness and color) as key causal factors, while image structure and semantic content are non-causal factors that should remain invariant.

Based on Pearl's causal intervention theory~\cite{pearl2009causality}, we implement three different causal interventions on causal factors. In an ideal latent space, low-light images after perfect causal intervention should be consistent with ground-truth images in their representation. Therefore, we use ground-truth images as the only positive samples while treating various intervention results as negative samples, guiding the encoder to achieve causal alignment through ground-truth supervision.

Meanwhile, inspired by~\cite{xu2024causality, hu2024interpreting} regarding the definition of causal principles, we formulate the causal intervention principles for LLIE:
\renewcommand{\theenumi}{\Roman{enumi}} 

(\uppercase\expandafter{\romannumeral1}) \textbf{Intervention Effectiveness}: Interventions should target low-frequency degradation causal factors. In the context of LLIE, this means focusing on features related to illumination and color, rather than on image-specific content, scene information, or other non-causal information.

(\uppercase\expandafter{\romannumeral2}) \textbf{Intervention Harmlessness}: An important principle of causal intervention is to ensure non-causal factors remain unchanged while intervening on causal factors. Therefore, we strive to maintain the scene semantic information consistency of the post-intervention image.  

\begin{figure}[t]
    \centering
    \includegraphics[width=0.9\linewidth]{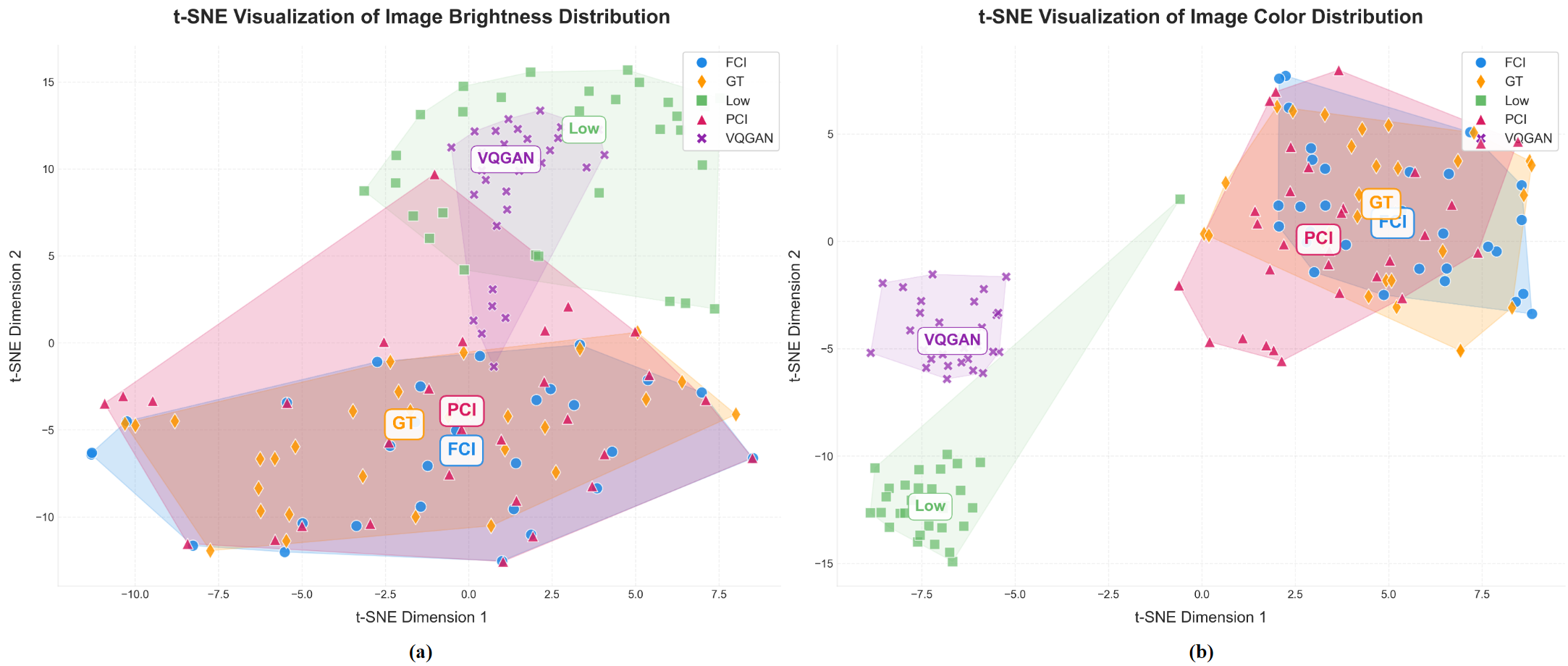}
    \vspace{-0.2cm}
    \caption{t-SNE~\cite{van2008visualizing} visualizations of brightness (a) and color (b) distributions using the LSRW-Huawei dataset. Each point represents different image outputs or inputs: Low denotes original low-light images; GT represents normal-light ground truth images; VQGAN shows outputs from a VQGAN model pre-trained on high-quality normal-illumination images; PCI refers to outputs from our model utilizing only the Pixel-level Causal Intervention module; and FCI includes outputs from our complete architecture with both PCI and Feature-level Causal Intervention modules.}
    \vspace{-0.2cm}
    \label{fig:t-sne}
\end{figure}

\begin{figure*}[t]
    \centering
    \includegraphics[width=1\linewidth]{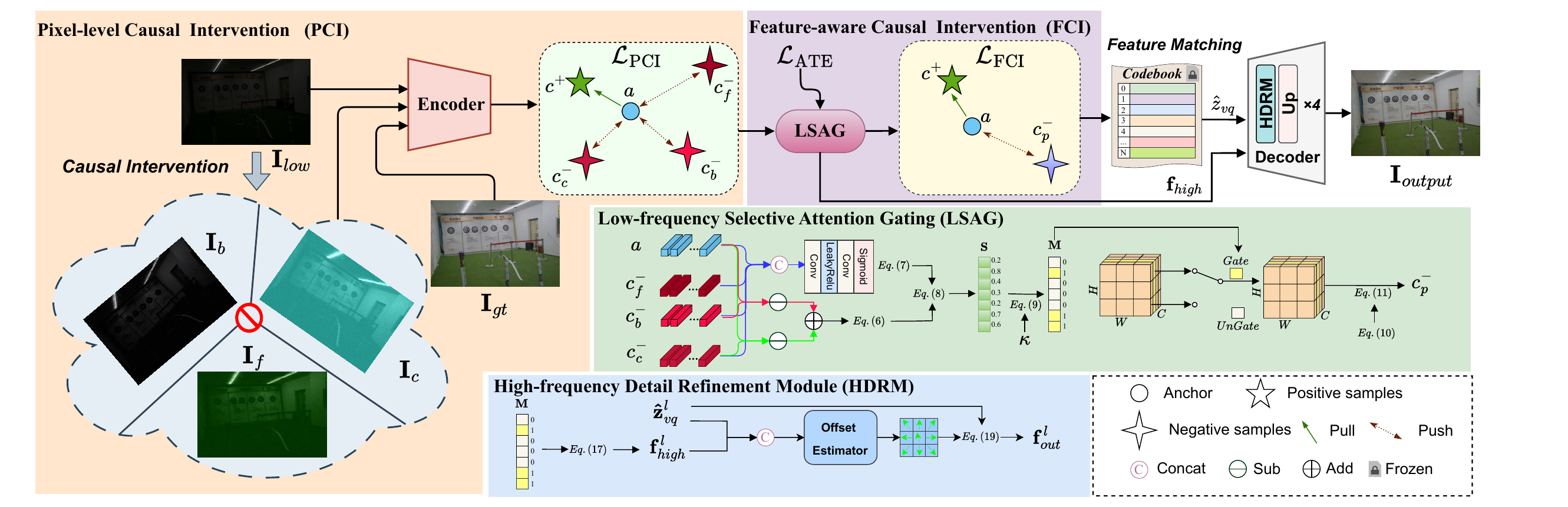}
    \vspace{-0.5cm}
    \caption{Overall architecture of our proposed CIVQLLIE.}
    \label{fig:network}
\end{figure*}

\subsection{Overall Architecture}  
\label{sec:overview} 
We first establish a high-quality vector quantization (VQ) codebook as a prior for our low-light image enhancement framework by pre-training a VQGAN architecture~\cite{esser2021taming} on the DIV2K~\cite{agustsson2017ntire} and Flickr2K~\cite{lim2017enhanced} high-quality image datasets. The vector quantization codebook $\mathcal{Z}$ serves as a high-quality prior containing numerous visual cues. However, due to severe illumination degradation and distribution shift, low-light images typically fail to activate the correct tokens in this pre-trained codebook, as shown in Fig.\ref{fig:t-sne}.
Through observations, we draw three key conclusions: (1) VQGAN outputs significantly deviate from the GT distribution and remain closer to the Low distribution, demonstrating that low-light images fail to properly activate the correct tokens in the pre-trained codebook; (2) After incorporating PCI, the output distribution shifts toward GT, though some samples still exhibit characteristics of the Low distribution; (3) When both PCI and FCI are applied, the output distribution aligns more closely with GT, validating that our proposed causal intervention methods effectively activate the appropriate tokens in the codebook. 

Fig.~\ref{fig:network} illustrates the architecture of our proposed CIVQLLIE framework. The pipeline consists of four key components: Pixel-level Causal Intervention (PCI), Feature-aware Causal Intervention (FCI), Codebook Matching, and Detail-preserving Decoder. The low-light input image $\mathbf{I}_{low}$ undergoes causal interventions to generate intervention images ($\mathbf{I}_b$, $\mathbf{I}_c$, and $\mathbf{I}_{f}$). Embedded features are then extracted through a unified trainable encoder: anchor $a$, positive sample ${c}^{+}$, and interfering negative samples ${c}^{-}_{b}$, ${c}^{-}_{c}$, and ${c}^{-}_{f}$. Causal metric learning is performed via $\mathcal{L}_{PCI}$.  
The FCI module employs Low-frequency Selective Attention Gating (LSAG) to identify and intervene on illumination-sensitive channels, combined with $\mathcal{L}_{ATE}$ to simultaneously adhere to causal principles, generating perturbed negative sample ${c}^{-}_{p}$. These perturbed features undergo causal consistency measurement through $\mathcal{L}_{FCI}$ before being matched with the pre-trained high-quality vector quantization codebook $\mathcal{Z}$ to obtain $\hat{z}_{vq}$. Finally, the decoder processes these quantized features through cascaded High-frequency Detail Refinement Modules (HDRM) and upsampling operations to produce the enhanced output image $\mathbf{I}_{output}$. HDRM leverages high-frequency channel information in deformable convolutions to adaptively refine fine-grained details.  
For detailed information, please refer to our supplementary material.

\subsection{Pixel-level Causal Intervention (PCI)}  
As shown in Fig.\ref{fig:t-sne}, low-light images exhibit significant distribution shifts compared to normal-light images, hindering effective codebook token matching. Since low-light degradation primarily affects visual content through illumination and color distortion (low-frequency components), referring to causal intervention principle  (\uppercase\expandafter{\romannumeral1}), the most direct approach is to perform causal intervention directly at the pixel level. Thus, as shown in Fig.~\ref{fig:network}, we introduce a Pixel-level Causal Intervention (PCI) module that intervenes on causal factors in the YCrCb color space and frequency domain.

\noindent \textbf{Causal Intervention}: The PCI module implements three parallel strategies: brightness intervention, color intervention, and frequency domain fusion, each with learnable parameters $\boldsymbol{\theta}_b$, $\boldsymbol{\theta}_c$, and $\boldsymbol{\alpha}$.  
For brightness intervention, we convert $\mathbf{I}_{low}$ to YCrCb space and adjust the luminance channel $Y_{low}$ while preserving chrominance:
\begin{equation}  
Y_{i} = Y_{low} \cdot \boldsymbol{\theta}_b,  
\end{equation}  
while the color intervention modifies chrominance:
\begin{equation}  
[Cr_{i}, Cb_{i}] = [Cr_{low}, Cb_{low}] \cdot \boldsymbol{\theta}_c.  
\end{equation}  
Inspired by recent advances in frequency domain image processing~\cite{huang2022deep, wang2023fourllie}, we propose a hybrid frequency domain fusion mechanism that adaptively combines brightness and color interventions. First, we compute the amplitude and phase spectra of the two intervention results:  
\begin{equation}  
\begin{aligned}  
A_b, P_b &= |\mathcal{F}(\mathbf{I}_{b})|, \angle\mathcal{F}(\mathbf{I}_{b}), \\
A_c, P_c &= |\mathcal{F}(\mathbf{I}_{c})|, \angle\mathcal{F}(\mathbf{I}_{c}),  
\end{aligned}  
\end{equation}  
\noindent where $\mathcal{F}(\cdot)$ represents the two-dimensional Fourier transform, and $\mathbf{I}_b$ and $\mathbf{I}_c$ are the images after brightness and color intervention, respectively. We then adaptively fuse the amplitude spectra while preserving the phase information from the brightness intervention:  
\begin{equation}  
\begin{aligned}  
A_{f} &= \boldsymbol{\alpha} \cdot A_b + (1 - \boldsymbol{\alpha}) \cdot A_c, \\
\mathbf{I}_{f} &= \mathcal{F}^{-1}(A_{f} \cdot e^{j P_b}),  
\end{aligned}  
\end{equation}  
\noindent where $\mathcal{F}^{-1}(\cdot)$ represents the inverse Fourier transform. This decoupled intervention strategy aligns with human visual perception, as phase information largely determines the structural content of the image (non-causal factors), while amplitude modulation primarily affects perceptual qualities such as brightness and color richness (causal factors).  

\noindent  \textbf{Causal Intervention Metric.} After obtaining feature embeddings from the encoder, we define a causal intervention metric to optimize our encoder:  
\begin{equation}  
\begin{aligned}  
\resizebox{0.7\hsize}{!}{$  
\mathcal{L}_{PCI} = - \log \frac{\exp (\tau \cdot s^+)}{\exp (\tau \cdot s^+) + \sum_{i=b}^{N} \exp (\tau \cdot s_i^-)},$}  
\label{eq:l_pci}
\end{aligned}  
\end{equation}  
\noindent where $\tau$ represents the scaling factor, $s^+ = \text{sim}(a, \mathbf{c}^+)$, $s_i^- = \text{sim}(a, \mathbf{c}_i^-)$, and $\text{sim}$ is the cosine similarity metric between embeddings. Here, $N = \{b, c, f\}$, $i \in N$ represents respectively our interfering negative samples from brightness, color, and frequency interventions. $\mathcal{L}_{\text{PCI}}$ encourages the encoded low-light features $a$ to align with the ground truth normal-light features $\mathbf{c}^+$ (positive sample) while separating them from the features of the intervention images $\mathbf{c}_b^-$, $\mathbf{c}_c^-$, and $\mathbf{c}_f^-$ (negative samples). As seen in Fig.~\ref{fig:t-sne}, PCI effectively corrects the causal factors in the low-light image representation, thereby improving codebook matching in subsequent quantization processes. Notably, we did not consider causal intervention principle (\uppercase\expandafter{\romannumeral2}) because we believe that intervening on the low-frequency components of the original image at the pixel level does not alter the semantic consistency of the image.  

\subsection{Feature-level Causal Intervention (FCI)}  
As shown in Fig.\ref{fig:t-sne}, after PCI, some samples still deviate from the ground truth image distribution and remain closer to the low-light image distribution. Inspired by recent advances in latent space manipulation~\cite{xu2024causality}, we recognize that diverse implicit feature-level transformations in latent space can enhance the stability of feature matching and flexibly generalize to unknown domains compared to limited image-level transformations. Therefore, we propose a Feature-level Causal Intervention (FCI) module that operates directly in feature space. We leverage the intervention samples generated by PCI to identify and selectively modify the channels most sensitive to illumination (causal intervention principle (\uppercase\expandafter{\romannumeral1})), thereby further improving the accuracy of codebook token matching and enhancing generalization performance on real-world low-light images. FCI consists of three key components: Low-frequency Selective Attention Gating (LSAG), Average Treatment Effect (ATE)~\cite{angrist1995identification, vanderweele2013causal} semantic consistency analysis (causal intervention principle (\uppercase\expandafter{\romannumeral2})), and progressive causal intervention metric.  

\noindent \textbf{Low-frequency Selective Attention Gating (LSAG).} To adhere to causal intervention principle (\uppercase\expandafter{\romannumeral1}), our approach builds on the core insight that not all channels in the feature space are equally affected by illumination conditions; some channels are more susceptible to lighting changes, some are more sensitive to color, while others primarily encode semantic and structural information. Given this heterogeneity, we introduce a sensitivity analysis framework $\mathcal{S}$ to identify channels most affected by illumination degradation:  
\begin{equation}  
\mathbf{S} = \mathcal{S}(a,  {c}^{-}_{b}, {c}^{-}_{c}, {c}^{-}_{f}),  
\end{equation}  
\noindent where $\mathbf{S} \in [0,1]\in \mathbb{R}^{H \times W \times C}$ represents the channel sensitivity map, with higher values indicating greater sensitivity to illumination conditions. Here, $H, W, C$ are the dimensions of the embedded feature $a$. We implement sensitivity analysis through a combination of feature difference $\mathbf{S}_{d}\in \mathbb{R}^{H \times W \times C}$ and learned sensitivity estimation $\mathbf{S}_{l}\in \mathbb{R}^{H \times W \times C}$:  
\begin{equation}  
\begin{aligned}  
\mathbf{D}_{b} &= |a - {c}^{-}_{b}|, \quad \mathbf{D}_{c} = |a - {c}^{-}_{c}|, \\
\mathbf{S}_{d} &= \frac{\mathbf{D}_{b} + \mathbf{D}_{c}}{\max(\mathbf{D}_{b} + \mathbf{D}_{c}) + \epsilon},  
\end{aligned}  
\end{equation}  
\begin{equation}  
\mathbf{S}_{l} = \Phi_S([a,  {c}^{-}_{b}, {c}^{-}_{c}, {c}^{-}_{f}]),  
\end{equation}  
\noindent where $\mathbf{D}_{b}\in \mathbb{R}^{H \times W \times C}$ and $\mathbf{D}_{c}\in \mathbb{R}^{H \times W \times C}$ represent the differences between $a$ after PCI processing and the causal intervention samples, $\mathbf{S}_{d}$ is the result after normalization, $\epsilon$ is a negligibly small value to prevent division by zero, $\Phi_S(\cdot)$ represents a lightweight sensitivity estimation network composed of two convolutional layers and two activation functions. After determining the low-frequency sensitivity estimate $\mathbf{S}= \mathbf{S}_{l} +\mathbf{S}_{d}$, we obtain the mask index $\mathbf{M}$ for channels sensitive to illumination and color through a binary mask:
\begin{equation}  
\mathbf{M} = (\mathbf{S} > \kappa),  
\end{equation}  
\noindent where $\kappa$ is a learnable threshold parameter initialized to 0.5, used to determine which channels need intervention. Our feature-level causal intervention strategy consists of two parts. First, we generate adaptive perturbations $\mathbf{p}\in \mathbb{R}^{H \times W \times C}$ through a lightweight feature modulation network $\Phi_P$ (composed of two convolutional layers with batch normalization and LeakyReLU activation functions):  
\begin{equation}  
\mathbf{p} = \Phi_P([a,  {c}^{-}_{b}, {c}^{-}_{c}, {c}^{-}_{f}]).  
\end{equation}  
Second, we control the intervention magnitude through a learnable intervention strength parameter:  
\begin{equation}  
{c}^{-}_{p} = a \cdot \mathbf{M} \cdot \mathbf{p}.
\end{equation}

\begin{table*}[ht]  
\centering  
\resizebox{\textwidth}{!}{  
\begin{tabular}{lcccccccccc}  
\toprule  
\multirow{2}{*}{\textbf{Methods}} & \multicolumn{2}{c}{\textbf{LOL-v1}} & \multicolumn{2}{c}{\textbf{LOL-v2-Real}} & \multicolumn{2}{c}{\textbf{LOL-v2-Syn}} & \multicolumn{2}{c}{\textbf{LSRW-Huawei}} & \multicolumn{2}{c}{\textbf{LSRW-Nikon}} \\   
\cmidrule(lr){2-3} \cmidrule(lr){4-5} \cmidrule(lr){6-7} \cmidrule(lr){8-9} \cmidrule(lr){10-11}  
& PSNR ↑ & SSIM ↑ & PSNR ↑ & SSIM ↑ & PSNR ↑ & SSIM ↑ & PSNR ↑ & SSIM ↑ & PSNR ↑ & SSIM ↑ \\
\midrule     
Kind~\cite{kind} (MM, 19) & 20.87 & 0.7995 & 17.54 & 0.6695 & 22.62 & 0.9041 & 16.58 & 0.5690 & 11.52 & 0.3827 \\   
MIRNet~\cite{lowlight9} (ECCV, 20) & \underline{24.14} & 0.8305 & 22.11 & 0.7942 & 22.52 & 0.8997 & 19.98 & 0.6085 & 17.10 & 0.5022 \\   
Kind++~\cite{kind++} (IJCV, 21) & 18.97 & 0.8042 & 19.08 & 0.8176 & 21.17 & 0.8814 & 15.43 & 0.5695 & 14.79 & 0.4749 \\    
SNR-Aware~\cite{lowlight8} (CVPR, 22) & 23.93 & 0.8460 & 21.48 & 0.8478 & 24.13 & 0.9269 & 20.67 & 0.5911 & 17.54 & 0.4822 \\   
FourLLIE~\cite{wang2023fourllie} (ACMMM, 23) & 20.99 & 0.8071 & 23.45 & 0.8450 & 24.65 & 0.9192 & 21.11 & 0.6256 & 17.82 & 0.5036 \\   
SNR-SKF~\cite{wu2023learning} (CVPR, 23) & 20.51 & 0.7110 & 21.82 & 0.7468 & 17.21 & 0.7738 & 16.21 & 0.5560 & 16.53 & 0.4415 \\   
UHDFour~\cite{li2023embedding} (ICLR, 23) & 22.89 & 0.8147 & 27.27 & 0.8579 & 23.64 & 0.8998 & 19.39 & 0.6006 & \underline{17.94} & 0.5195 \\   
Retinexformer~\cite{cai2023retinexformer} (ICCV, 23) & 22.71 & 0.8177 & 24.55 & 0.8434 & 25.67 & 0.9295 & 21.23 & 0.6309 & 17.64 & 0.5082 \\    
DMFourLLIE \cite{zhang2024dmfourllie} (ACMMM, 24) & 22.98 & 0.8273 & 26.40 & 0.8765 & 25.83 & 0.9314 & 21.47 & 0.6331 & 17.04 & \underline{0.5294} \\  
UHDFormer \cite{wang2024uhdformer} (AAAI, 24) & 21.17 & 0.8218 & 19.71 & 0.8320  & 24.48 & 0.9277 & 20.64 & 0.6244 & 17.26 & 0.5230 \\  
Wave-Mamba \cite{zou2024wave} (ACMMM, 24) & 22.76 & 0.8419 & \underline{27.87} & \underline{0.8935} & 24.69 & 0.9271 & 21.19 & 0.6391 & 17.34 & 0.5192 \\    
RetinexMamba \cite{bai2024retinexmamba} (Arxiv, 24) & 23.15 & 0.8210 & 27.31 & 0.8667 & \underline{25.89} & 0.9351 & 20.88 & 0.6298 & 17.59 & 0.5133 \\  
CWNet \cite{zhang2025cwnet} (ICCV, 25) & 23.81 & 0.8490 & 27.39 & \textbf{0.9005} & 25.74 & 0.9365 & \underline{21.50} & \underline{0.6397} & 17.38 & 0.5119 \\  
CIDNet \cite{bai2024retinexmamba} (CVPR, 25) & \underline{23.83} & \underline{0.8574} & - & - & 25.70 & \underline{0.9419} & 20.30 & 0.6054& 17.16 & 0.4975 \\  
\midrule   
CIVQLLIE  (Ours) & \textbf{24.79} & \textbf{0.8871} & \textbf{28.67} & 0.8783 & \textbf{26.15} & \textbf{0.9471} & \textbf{21.82} & \textbf{0.6543} & \textbf{18.01} & \textbf{0.5357} \\  
\bottomrule  
\end{tabular}  }  
\vspace{-0.3cm}
\caption{Quantitative comparison on LOL-v1, LOL-v2-Real, LOL-v2-Syn, LSRW-Huawei, and LSRW-Nikon. The best and second-best results are highlighted in bold and underlined, respectively. All methods did not use the GT-Mean strategy.}  
\vspace{-0.2cm}
\label{tab:com}  
\end{table*}  

\noindent  \textbf{ATE Semantic Consistency Analysis.}  
Average Treatment Effect (ATE)~\cite{angrist1995identification, vanderweele2013causal} is particularly important in causal inference research as it quantifies causal effects by representing the difference between average outcomes in treatment and control groups. It provides the average causal effect of treatment across the entire population, beyond the individual level. This helps mitigate individual differences caused by interventions, leading to a clearer understanding of the overall treatment effect.

Since feature-level intervention is a double-edged sword that inevitably causes information loss, especially at the feature level, to ensure the invariance of non-causal factors and follow the ATE and causal intervention principle (\uppercase\expandafter{\romannumeral2}), we adopt the Average Treatment Effect (ATE) principle from causal inference in CEM~\cite{hu2024interpreting} to quantify the expected difference in semantic representations before and after intervention:  
\begin{equation}  
\phi_{\mathcal{F}}[{c}^{-}_{p}] = \text{ATE}_{\mathcal{F}}[{c}^{-}_{p}] =   
\mathcal{M}_R(a) - \mathcal{M}_R({c}^{-}_{p}),  
\end{equation} 
\begin{equation}  
\mathcal{L}_{ATE} = \arg\min\phi_{\mathcal{F}}[{c}^{-}_{p}] .  
\end{equation}  
\noindent where $\mathcal{M}_R(\cdot)$ is pre-trained VGG16~\cite{simonyan2014very}. This metric directly compares the difference in semantic space between low-light images and their intervened counterparts, evaluating the effectiveness of interventions in preserving image semantic content. By minimizing this difference, we ensure that core semantic information is maintained during the enhancement process while effectively improving low-frequency visual features.  

\noindent \textbf{Progressive Causal Intervention Metric.} Similar to $\mathcal{L}_{PCI}$, we encourage the learning objective to align more closely with the normal-light distribution while maintaining a certain distance from the intervention features:  
\begin{equation}  
\begin{aligned}  
\resizebox{0.7\hsize}{!}{$  
\mathcal{L}_{FCI} = - \log \frac{\exp (\tau \cdot s^+)}{\exp (\tau \cdot s^+) +  \exp (\tau \cdot s^-)},$}  
\end{aligned}  
\end{equation}    
where $\tau$ and  $s^+$ refer to Eq.~\ref{eq:l_pci}, $s^- = \text{sim}(a, \mathbf{c}_p^-).$
This measurement objective ensures that our interventions effectively bridge the distribution gap between low-light and normal-light features, thereby promoting more accurate codebook token matching. The complete FCI process can be summarized as:  
\begin{align}  
{c}^{-}_{p}, \mathbf{S}, \mathbf{M} &= \text{LSAG}(a,  {c}^{-}_{b}, {c}^{-}_{c}, {c}^{-}_{f}),\\
\mathcal{L}_{FCI\text{-}t} &= \mathcal{L}_{FCI} + \lambda_{ate}\mathcal{L}_{ATE},   
\end{align}  
\noindent where $\lambda_{ate} = 0.5$ is weighting coefficient, the LSAG output includes not only perturbed features but also sensitivity maps and index masks, providing explainable insights into the intervention process by selectively modifying channels most sensitive to causal factors, adhering to causal principles and ATE semantic consistency. Our FCI effectively addresses feature-level distribution changes caused by low-light conditions, enabling more robust performance across various real-world scenarios  and facilitating the effective activation of the codebook prior. Finally, the anchor $a$ undergoes codebook discrete matching to obtain the quantized feature $\hat{z}_{vq}$. 

\begin{figure*}[t]
    \centering
    \includegraphics[width=1\linewidth]{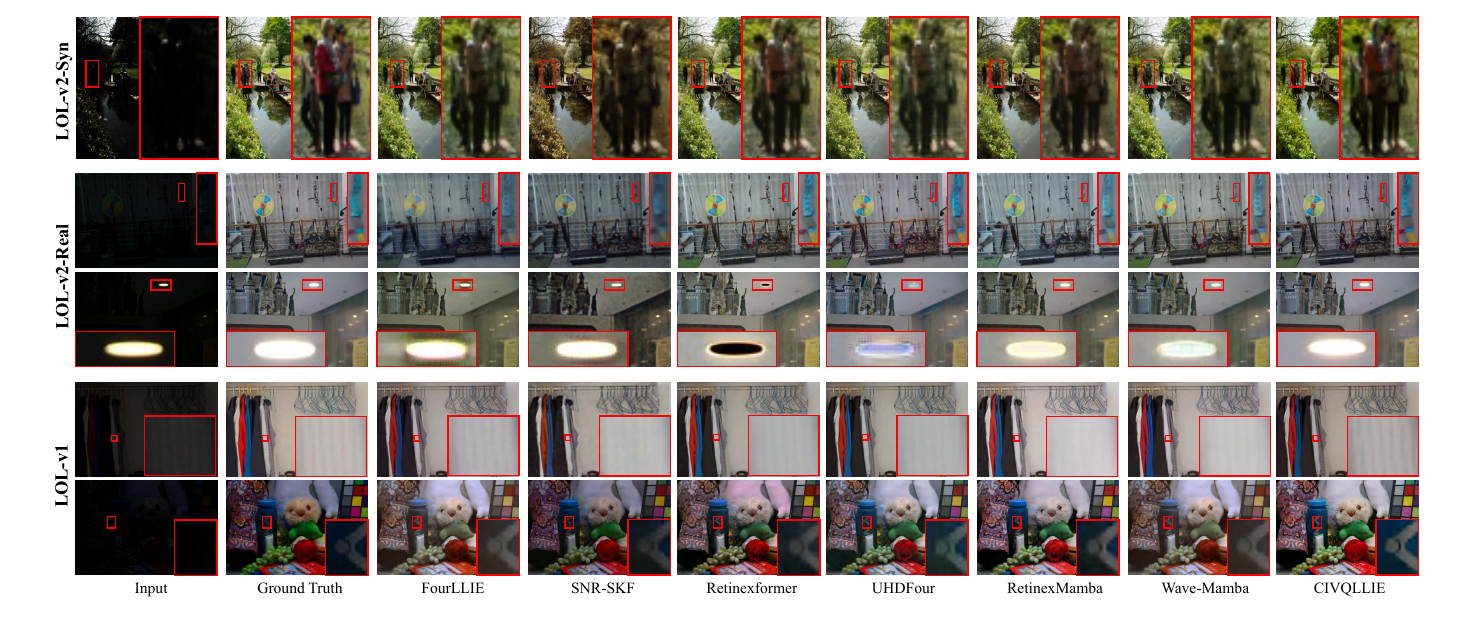}
    \vspace{-0.6cm}
    \caption{Visual comparison on LOL-v1, LOL-v2-Real, and LOL-v2-Syn datasets.}
    \vspace{-0.2cm}
    \label{fig:com_lol}
\end{figure*}

\begin{figure*}[t]
    \centering
    \includegraphics[width=1\linewidth]{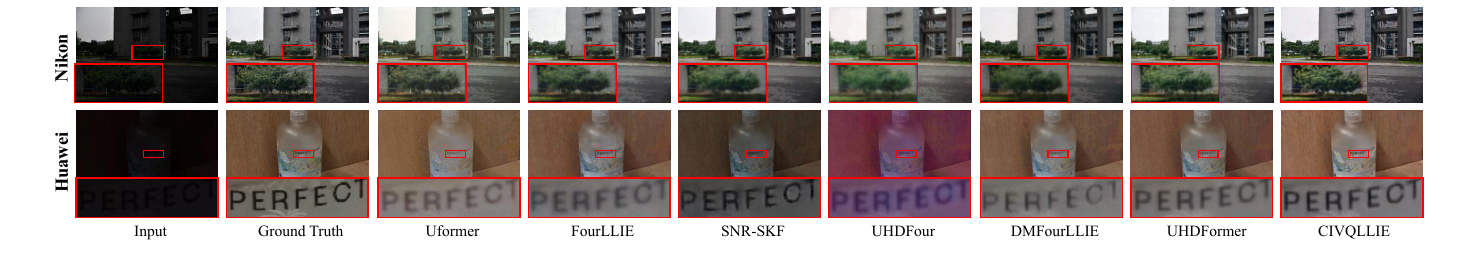}
    \vspace{-0.7cm}
    \caption{Visual comparison on LSRW-Huawei and LSRW-Nikon datasets.}
    \vspace{-0.2cm}
    \label{fig:com_lsrw}
\end{figure*}

\subsection{High-frequency Detail Refinement Module (HDRM)}  
\label{subsec:hdrm}  
 We consider the decoder as a refinement process aimed at restoring the image's details and structural information. Although VQGAN is highly effective in codebook matching at higher semantic levels, obtaining perfectly quantized features from the codebook remains challenging due to the inherent trade-off between illumination enhancement and texture preservation~\cite{zou2024vqcnir, wang2024perceplie}. To address this limitation, we introduce the High-frequency Detail Refinement Module (HDRM), which complements codebook representations with fine-grained details.  

As shown in the Fig.\ref{fig:network}, HDRM operates hierarchically from low to high resolution, progressively refining features by fusing matched high-quality codebook features with features from high-frequency sensitive channels. Specifically, utilizing the low-frequency sensitivity mask $\mathbf{M}$ obtained from LSAG, we recognize that high-frequency components encoding texture details, semantic boundaries, and structural information are typically preserved in the standard decoding process. Therefore, we reversely derive the features $\mathbf{f}_{high}$ that primarily encode high-frequency channels:  
\begin{equation}  
\mathbf{f}_{high} = a \odot (1-\mathbf{M}), 
\end{equation}   
\noindent where $\mathbf{f}_{high}$ is subsequently processed through a $1 \times 1$ convolution to match the dimensionality of $\hat{z}_{vq}$. At each resolution level $l$, we generate spatially adaptive sampling offsets and apply deformable convolution to enhance feature refinement:   
\begin{align}  
\mathbf{f}^l_{offset} &= \Psi_{offset}(\text{Concat}(\mathbf{\hat{z}}^l_{vq}, \mathbf{f}^l_{high})), \\
\mathbf{f}^l_{out} &= \text{DeformConv}(\mathbf{\hat{z}}^l_{vq}, \mathbf{f}^l_{offset}),  
\end{align}  
\noindent where $\Psi_{offset}$ represents the offset prediction network, and DeformConv represents deformable convolution~\cite{dai2017deformable}. Unlike standard convolutions with fixed receptive fields, deformable convolution dynamically adjusts sampling positions based on learned offsets, enabling content-adaptive feature aggregation. This cascaded offset field prediction allows the model to maintain consistency between different feature levels while allowing adaptive spatial transformations. By incorporating deformable sampling for spatial feature alignment, our approach can more accurately preserve image details that might be lost during the quantization process.

\section{Experiments}
\label{experiments}
\subsection{Datasets and Implementation Details}  
\textbf{Datasets.}
We evaluate on several LLIE benchmarks. The training datasets include four paired datasets with the following counts: LOL-v1~\cite{lol} (485 pairs), LOL-v2-Synthetic (900 pairs), LSRW-Huawei~\cite{lsrw} (2,450 pairs), and LSRW-Nikon (2,450 pairs). For evaluation, we use LOL-v2-Real (100 pairs) and five unpaired datasets: DICM~\cite{dicm} (64 images), LIME~\cite{lime} (10 images), MEF~\cite{mefl} (17 images), NPE~\cite{npe} (85 images), and VV (24 images). 

\noindent \textbf{Implementation Details.}  
VQGAN backbone is pre-trained on DIV2K and Flickr2K for $3.5 \times 10^5$ iterations to establish a robust codebook. CIVQLLIE is implemented in PyTorch and trained end-to-end. We apply random cropping to $256 \times 256$ patches and horizontal flips for augmentation. The Adam optimizer is used ($\beta_1 = 0.9$, $\beta_2 = 0.99$) with a fixed learning rate of $1 \times 10^{-4}$, training for $2.0 \times 10^5$ iterations with a batch size of 8 on two NVIDIA RTX 4090 GPUs.

\noindent \textbf{Comparison with State-of-the-Art Methods.}  
We benchmark against SOTA approaches from multiple categories: CNN-based methods (Kind~\cite{kind}, Kind++~\cite{kind++}, MIRNet~\cite{lowlight9}, SGM~\cite{lol}, HDMNet~\cite{liang2022learning}, SNR-Aware~\cite{lowlight8}, SNR-SKF~\cite{wu2023learning}, CIDNet~\cite{yan2025hvi}), Transformer-based methods (Uformer~\cite{wang2022uformer}, Retinexformer~\cite{cai2023retinexformer}, UHDFormer~\cite{wang2024uhdformer}), Fourier-based methods (FourLLIE~\cite{wang2023fourllie}, UHDFour~\cite{UHDFourICLR2023}, DMFourLLIE~\cite{zhang2024dmfourllie}), and Mamba-based methods (RetinexMamba~\cite{bai2024retinexmamba}, Wave-Mamba~\cite{zou2024wave}, CWNet~\cite{zhang2025cwnet}). All models are trained on identical datasets using publicly available implementations.  

\noindent \subsection{Comparison with Current Methods}
\textbf{Quantitative comparison on paired datasets.} We report PSNR and SSIM~\cite{SSIM} metrics. As shown in Tab.~\ref{tab:com}, CIVQLLIE consistently outperforms state-of-the-art methods across almost all datasets.  

\noindent \textbf{Visual comparisons.} Fig.~\ref{fig:com_lol} validates our quantitative results. On LOL-v2-Syn, our method achieves optimal color and brightness. For LOL-v2-Real, we demonstrate superior detail and color rendering in the first row, while the second row reveals brightness inconsistencies in Retinexformer and UHDFour, with our results appearing the most natural and closest to ground truth. On LOL-v1, we excel in texture recovery, particularly in the second row, where our method surpasses others in color, brightness.
Fig.~\ref{fig:com_lsrw} showcases our method's effectiveness on smartphone-captured datasets. In the first row, the magnified region highlights our approach's natural tree coloration and clear structural boundaries, unlike competing methods that either under-enhance dark regions or introduce artifacts. In the second row, we effectively preserve structural details and brightness, while others falter in detail preservation and color consistency. \textbf{More results are provided in the supplementary material.}  

\noindent \textbf{Efficiency Comparison.}  
We conducted an efficiency comparison with recent VQGAN-based LLIE methods. As shown in Tab.\ref{tab:efficiency_comparison}, our method strikes a favorable balance between efficiency, parameter count, and performance.

\begin{table}[t]  
    \centering  
    \resizebox{1.0\columnwidth}{!}{ 
    \begin{tabular}{lcccc}  
        \toprule  
        \multirow{2}{*}{\textbf{Setting}} &   
        \multicolumn{2}{c}{\textbf{LOL-v1}} &   
        \multicolumn{2}{c}{\textbf{LSRW-Huawei}} \\
         & \textbf{PSNR ↑} & \textbf{SSIM ↑} & \textbf{PSNR ↑} & \textbf{SSIM ↑} \\
        \midrule  
        Baseline & 23.16 & 0.8653 & 21.06 & 0.6428 \\
        w/o PCI ($\mathcal{L}_{PCI}$) & 23.47 & 0.8740 & 21.35 & 0.6441 \\
        w/o FCI ($\mathcal{L}_{FCI}$) & 23.26 & 0.8597 & 21.12 & 0.6428 \\
        w/o LSAG & 24.03 & 0.8745 & 21.52 & 0.6489 \\
        w/o $\mathcal{L}_{ATE}$ & 24.73 & 0.8815 & 21.74 & \textbf{0.6561} \\
        w/o HDRM & 24.40 & 0.8781 & 21.52 & 0.6498 \\ \hline
        CIVQLLIE (Full Model) & \textbf{24.79} & \textbf{0.8871} & \textbf{21.82} & 0.6543 \\
        \bottomrule  
    \end{tabular}  }
    \vspace{-0.2cm}
    \caption{Ablation study. Baseline is the performance with the encoder-decoder and codebook structure fine-tuned on LOL-v1 and LSRW-Huawei datasets.} 
    \label{table:ablation}  
    \vspace{-0.3cm}
\end{table}

\begin{figure}[t]
    \centering
    \includegraphics[width=1\linewidth]{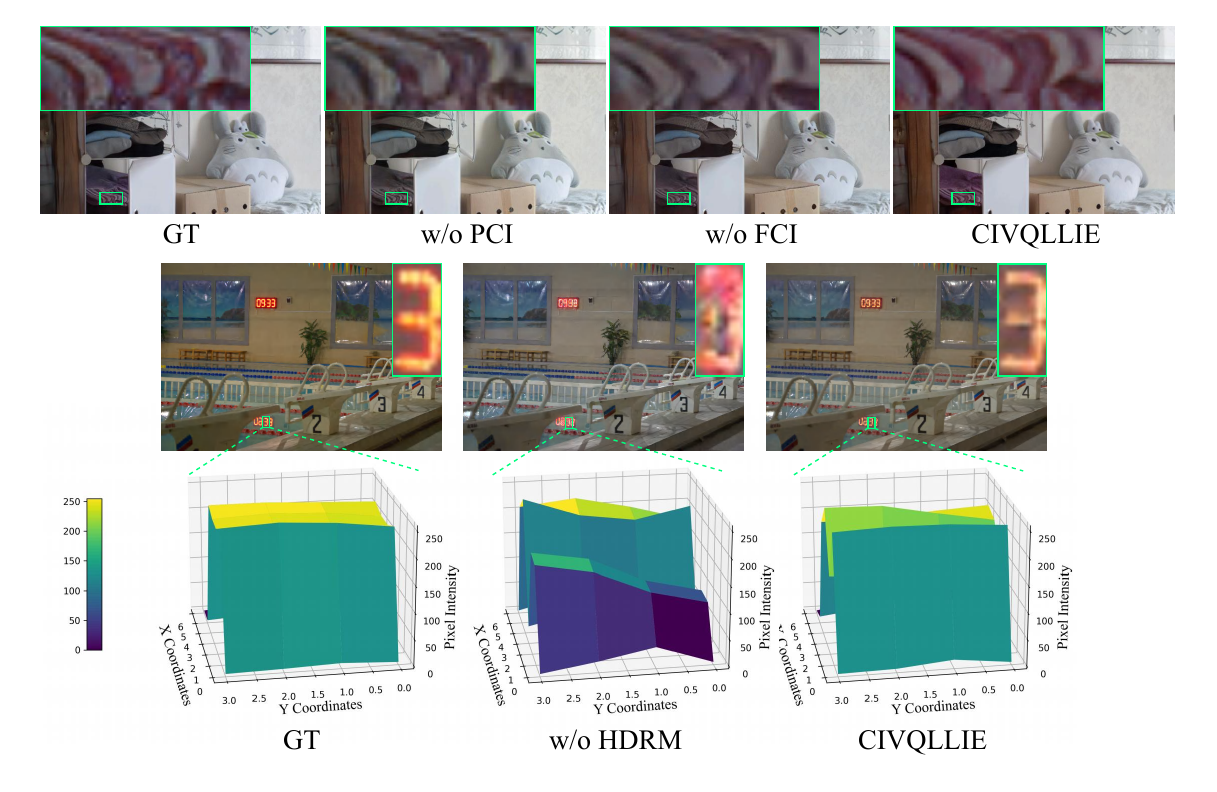}
    \vspace{-0.4cm}
    \caption{Ablation visualization comparison of PCI, FCI, and HDRM. In the second row, we perform a gradient map visualization comparison on the magnified area.}
    \label{fig:ab_show}
\end{figure}

\begin{table}[t]  
    \centering  
    \resizebox{1.0\columnwidth}{!}{ 
    \begin{tabular}{lcccccc}  
        \toprule  
        \textbf{Method} & \textbf{Parameters (M)} & \textbf{GFLOPs} & \textbf{Time (s)} & \textbf{PSNR} & \textbf{SSIM} \\  
        \midrule  
        GLARE (ECCV'24) & 71.72 & 1091.84 & 0.349 & 23.34 & 0.8430 \\  
        VQCNIR (AAAI'24) & \textbf{45.90} & \underline{325.27} & \underline{0.247} & \underline{23.47} & \underline{0.8482} \\  
        \midrule  
        \textbf{Ours} & \underline{59.47} & \textbf{245.45} & \textbf{0.195} & \textbf{24.79} & \textbf{0.8871} \\  
        \bottomrule  
    \end{tabular}  
    }
    \vspace{-0.1cm}
    \caption{Computational comparison with VQGAN-based methods on LOL-v1 dataset. Our method balances efficiency, parameter count, and performance effectively.}  
    \label{tab:efficiency_comparison}  
\end{table}

\subsection{Ablation studies}
We conducted ablation experiments to evaluate the contributions of each core component, as shown in Tab.~\ref{table:ablation}. Independent ablations confirm that all configurations enhance the framework's performance, with the full model yielding the best results.
Fig.~\ref{fig:ab_show} visualizes the ablation results. In the first row, removing PCI and FCI results in reduced brightness and low-frequency color representation, while the complete CIVQLLIE shows more natural and vibrant outputs, underscoring the importance of PCI and FCI in handling low-frequency causal factors. The second row visualizes edge gradients, emphasizing HDRM's effectiveness in refining spatial details and enhancing image sharpness.

\section{Conclusion and Limitations}
\label{conlusion}
\textbf{Conlusion:} We propose a novel framework for LLIE that incorporates causal reasoning into vector quantization. Our method systematically addresses distribution shifts in degraded inputs through a multi-level causal intervention strategy, including PCI and FCI. Additionally, HDRM refines structural details, enabling superior enhancement performance on both standard benchmarks and real-world low-light images.

\textbf{Limitations:}
Despite the effectiveness of PCI and FCI in addressing low-frequency causal factors, there remains significant potential to explore the degree and methods of intervention. For example, implementing strategies for strong and weak interventions could enhance the granularity of interventions within the causal space. Additionally, integrating interventions from other modalities, such as language-based inputs, or exploring cross-modal strategies like style transfer without altering semantics, could open new research avenues. These points suggest the need for further exploration of causal interventions in improving LLIE techniques.

\bibliography{aaai2026}



\setcounter{section}{0}
\renewcommand{\thesection}{\Alph{section}}  
\clearpage
\setcounter{page}{1}

\twocolumn[ 
    \begin{center}
        \textbf{\Large Supplementary Material}
    \end{center}
    \vspace{1em}  
]
\setcounter{figure}{0}
\setcounter{table}{0}

\section{Related Work}

\subsection{Low-Light Image Enhancement}  

Early methods like histogram equalization~\cite{bovik2010handbook,pisano1998contrast} and Retinex theory~\cite{land1977retinex,ng2011total} improved visibility through intensity redistribution or illumination-reflectance decomposition. However, these approaches relied on simplified assumptions that failed under complex lighting conditions.  
Deep learning has through several development stages. Initial CNN-based methods such as LLNet~\cite{lore2017llnet} learned direct mappings from low-light to normal-light images. Subsequently, physics-informed approaches integrated Retinex theory into learning frameworks, including Deep Retinex Decomposition~\cite{wei1808deep}, KinD~\cite{zhang2019kindling}, and more recently, Retinexformer~\cite{cai2023retinexformer} and Retinexmamba~\cite{bai2024retinexmamba}. These methods enhanced interpretability but remained constrained by their simplified physical assumptions.  
Recent advances have explored novel solution spaces: (1) Frequency-domain methods like FourLLIE~\cite{wang2023fourllie} and DMFourLLIE~\cite{zhang2024dmfourllie} manipulate amplitude components in Fourier space; (2) Transformer-based models~\cite{cai2023retinexformer,wang2024uhdformer} address long-range dependencies in lighting conditions, with UHDformer~\cite{wang2024uhdformer} specifically targeting ultra-high-definition restoration; and (3) State space models (SSMs) offering efficient alternatives to transformers, as demonstrated by Wave-Mamba~\cite{zou2024wave} and LLEMamba~\cite{zhang2024llemamba}, which balance global context modeling with local detail preservation.  

Despite these advances, most existing methods~\cite{zhang2022deep,huang2022deep,zhang2024dmfourllie,wu2023learning,zhang2024adapt,bai2024retinexmamba,feijoo2025darkir,yan2025hvi} suffer from fundamental limitations: they either lack interpretability (end-to-end approaches), struggle with generalization in extreme conditions (domain shift issues), or rely on potentially unreliable priors derived from already degraded inputs. 
To address these challenges, we propose a novel approach leveraging vector quantization (VQ) codebooks as implicit priors, eliminating dependence on handcrafted or estimated priors while providing a generalizable, high-quality feature repository for robust low-light image enhancement across diverse scenarios. 

\begin{table*}[t]  
\centering  
\resizebox{0.8\textwidth}{!}{  
\begin{tabular}{l|ccccc|c}  
\toprule  
\textbf{Methods} & \textbf{LIME} & \textbf{VV} & \textbf{DICM} & \textbf{NPE} & \textbf{MEF} & \textbf{AVG} \\
\midrule  
GLARE~\cite{zhou2024glare} & 4.343 & 4.262 & 4.858 & 5.467 & 3.906 & 4.567 \\
LYTNet~\cite{brateanu2401lyt} & 4.540 & 4.639 & 4.187 & 4.372 & 3.959 & 4.339 \\
LANet~\cite{yang2023learning} & 3.680 & 4.670 & 3.902 & 4.856 & 3.931 & 4.208 \\
FourLLIE~\cite{wang2023fourllie} & 3.814 & 4.222 & 4.365 & 4.914 & 3.831 & 4.229 \\
Wave-Mamba~\cite{zou2024wave} & 4.451 & 4.713 & 4.568 & 4.545 & 4.767 & 4.608 \\
Retinexformer~\cite{cai2023retinexformer} & 3.442 & 3.719 & 4.013 & 3.891 & 3.730 & 3.759 \\
UHDFormer~\cite{wang2024uhdformer} & 4.352 & 4.287 & 4.424 & 4.406 & 4.745 & 4.442 \\
FECNet~\cite{huang2022deep} & 6.043 & 3.751 & 4.140 & 4.508 & 4.716 & 4.632 \\
CWNet~\cite{zhang2025cwnet} & 3.921 & 3.583 & 3.662 & 3.610 & 3.741 & 3.701 \\  
\midrule  
\textbf{Ours} & \textbf{3.179} & \textbf{2.694} & \textbf{3.486} & \textbf{3.443} & \textbf{3.052} & \textbf{3.170} \\
\bottomrule  
\end{tabular}  }
\caption{Quantitative comparison using NIQE metric (lower is better) on unpaired real-world datasets.} 
\label{tab:niqe} 
\end{table*}

\subsection{Discrete Codebook Learning and Causal Analysis}  

\textbf{Discrete Codebook Learning.} Vector quantization has emerged as a powerful technique for learning discrete representations in generative models. VQ-VAE~\cite{van2017neural} pioneered this approach by learning a discrete latent codebook for high-fidelity image reconstruction, which VQGAN~\cite{esser2021taming} further enhanced through adversarial supervision to improve perceptual quality and generation diversity. These advances have led to successful applications in various low-level vision tasks, including super-resolution~\cite{chen2022real}, image dehazing~\cite{fu2025iterative}, and low-light enhancement~\cite{liu2023low}.  

Despite these advances, a critical challenge persists in LLIE: degraded low-light features often differ significantly from those stored in pre-trained codebooks derived from high-quality images. This misalignment hinders effective codebook token activation, resulting in suboptimal restoration with amplified artifacts. Our work addresses this limitation through causal interventions that systematically reduce the feature gap between degraded inputs and high-quality codebook representations, enabling precise feature realignment while maintaining flexibility for diverse restoration scenarios.  

\subsection{Causal Analysis in Computer Vision.} Causal inference has demonstrated significant potential across various computer vision domains, including domain generalization~\cite{xu2024causality}, metric learning~\cite{yan2024causality}, and modality alignment~\cite{li2023towards, li2024multimodal}. However, its application to image restoration remains largely unexplored. While network interpretability in low-level vision has gained increasing attention~\cite{gu2021interpreting,gu2023networks}, most research has focused on correlation studies rather than comprehensive examinations of causal relationships. The work of~\cite{hu2024interpreting} represents the first attempt to introduce causal analysis for interpretability in low-level vision, though it primarily focused on explaining model decisions through causal effect maps rather than leveraging causality for performance enhancement.

CWNet~\cite{zhang2025cwnet} is the first method that employs causal reasoning to guide Low-Light Image Enhancement (LLIE). By treating semantic factors as causal and color and brightness as non-causal, it ensures consistency of causal factors during the brightening process while eliminating interference from non-causal factors. In contrast, our analysis delves deeper, asserting that the directly relevant factors in LLIE should be brightness, which ought to be treated as a causal factor. By intervening on these causal factors to achieve enhancement, we argue that non-causal factors should remain unchanged during the brightening process, aligning more closely with Pearl~\cite{pearl2009causality}'s causal intervention theory. Furthermore, we enhance real-world performance by leveraging high-quality codebook priors.

\section{Vector Quantization (VQ) Codebook Prior Pre-training}
\begin{figure}[t]
    \centering
    \includegraphics[width=1\linewidth]{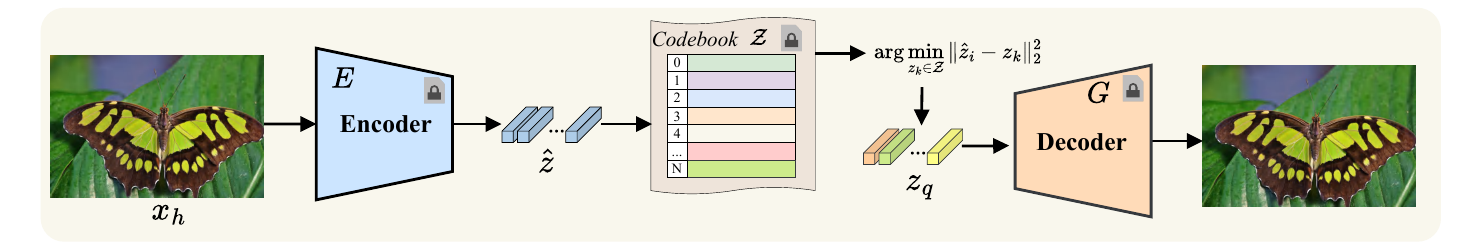}
    \caption{Overall framework of VQGAN. We first pre-train VQGAN to learn vector quantized codebook as high-quality prior information. }
    \label{fig:vqgan}
\end{figure}
We first establish a high-quality vector quantization (VQ) codebook as prior information for our low-light image enhancement framework. For this purpose, we build upon the VQGAN architecture~\cite{esser2021taming} as shown in Fig.~\ref{fig:vqgan}, which consists of an encoder $E$, a discrete codebook $\mathcal{Z}$, and a decoder $G$.

\textbf{VQGAN Structure :} In our experiments, we use a codebook with a size of \(1024\times512\). The input images have a size of \(256\times256\). In the encoding layer, the dimension is first increased through a convolutional layer. Then, it passes through a module consisting of four layers, each layer containing down-sampling and two standard residual blocks. The decoder is composed of a module with four layers, where each layer includes three standard residual blocks and up-sampling. Finally, the dimension is reduced through a convolutional layer to obtain the output image.

\textbf{Pre-training :} Given a high-quality normal-light image $x_h \in \mathbb{R}^{H \times W \times 3}$, the encoder maps it to a spatial latent representation $\hat{z} = E(x_h) \in \mathbb{R}^{h \times w \times n_z}$, where $n_z$ is the dimension of the latent vectors. Each element $\hat{z}_i \in \mathbb{R}^{n_z}$ of the latent representation is then quantized by finding its nearest neighbor in the codebook using Euclidean distance:  
\begin{equation}  
z_q = q(\hat{z}) := \left( \arg\min_{z_k \in \mathcal{Z}} \|\hat{z}_i - z_k\|_2^2 \right) \in \mathbb{R}^{h \times w \times n_z},  
\end{equation}  
where $z_q$ is the resulting discrete representation, $q(*)$ denotes the matching operation, and the codebook $\mathcal{Z} = \{z_k\}_{k=1}^K \subset \mathbb{R}^{K \times n_z}$ contains $K$ discrete codes. The decoder $G$ subsequently maps the quantized representation back to RGB space:  
\begin{equation}  
\hat{x}_h = G(z_q) \approx x_h.  
\end{equation}  

where $\hat{x}_h$ represents the reconstruction result. The VQGAN model and codebook are trained end-to-end using the following objective function: 
\begin{equation}  
\mathcal{L}_{VQ}(E, G, \mathcal{Z}) = \|x_h - \hat{x}_h\|_1 + \|\text{sg}[\hat{z}] - z_q\|_2^2 + \beta\|\hat{z} - \text{sg}[z_q]\|_2^2,  
\end{equation}  
where $\text{sg}[\cdot]$ represents the stop-gradient operation, and $\beta = 0.25$ is a hyperparameter balancing the commitment loss. The first term is the reconstruction loss, while the second and third terms are the codebook loss and commitment loss, respectively. With the pre-trained VQGAN, any high-quality normal-light image can be effectively encoded into and reconstructed from the discrete codebook space. This codebook captures standardized brightness and color patterns independent of illumination attenuation, providing strong prior information for low-light image enhancement.

\section{CIVQLLIE}
\textbf{CIVQLLIE Architecture: }CIVQLLIE builds upon the VQGAN framework with several key enhancements. We maintain the same encoder structure as VQGAN but optimize it through PCI and FCI modules to better align with the pre-trained codebook distribution. In the decoding stage, we augment the VQGAN decoder by incorporating HDRM layers before each standard layer, utilizing deformable convolutions to supplement fine-grained detail features missing from the codebook representation.  

\textbf{Optimization Objectives: }  
CIVQLLIE is trained with a composite loss function consisting of five components:  

\textbf{(1) Pixel Reconstruction Loss} measures the $\ell_1$ distance between the enhanced output $I_{output}$ and the ground truth $I_{gt}$:  
\begin{equation}  
\mathcal{L}_{pix} = \|I_{output} - I_{gt}\|_1.  
\end{equation}  

\textbf{(2) Codebook Matching Loss} enforces alignment between the quantized low-light features $\hat{z}_{vq}$ and the corresponding ground truth features $\mathbf{z}_{gt}$:  
\begin{equation}  
\mathcal{L}_{cp} = \|\hat{z}_{vq} - sg(\mathbf{z}_{gt})\|_2^2 + \beta \|sg(\hat{z}_{vq}) - \mathbf{z}_{gt}\|_2^2,  
\end{equation}  
where $sg(\cdot)$ denotes the stop-gradient operation and $\beta=0.25$ according to~\cite{esser2021taming}.  

\textbf{(3) Perceptual Loss} minimizes the feature-space distance extracted by VGG16~\cite{simonyan2014very} between the output and ground truth to enhance perceptual quality:  
\begin{equation}  
\mathcal{L}_{per} = \|\phi(I_{output}) - \phi(I_{gt})\|_2^2,  
\end{equation}  
where $\phi(\cdot)$ represents the features from the VGG16 network.  

\textbf{(4) Adversarial Loss} helps discover better feature matches during the codebook matching process, restoring realistic textures:  
\begin{equation}  
\mathcal{L}_{adv} = \log(1-D(I_{output})) + \log D(I_{gt}),  
\end{equation}  
where $D$ is a discriminator with the U-Net structure referred to in ~\cite{wang2021real}. 

\textbf{(5) Causal Intervention Loss} consists of two components:Pixel-level Causal Intervention (PCI) loss and Feature-level Causal Intervention (FCI) loss:  
\begin{equation}  
\mathcal{L}_{ca} = \mathcal{L}_{PCI} + \mathcal{L}_{FCI-t}.
\end{equation}

The total training loss for CIVQLLIE is:  
\begin{equation}  
\mathcal{L}_{t} = \lambda_{1}\mathcal{L}_{pix} + \lambda_{2}\mathcal{L}_{cp} + \lambda_{3}\mathcal{L}_{per} + \lambda_{4}\mathcal{L}_{adv} + \lambda_{5}(\mathcal{L}_{PCI} + \mathcal{L}_{FCI-t}),  
\end{equation}  
where $\lambda_{1}$, $\lambda_{2}$, $\lambda_{3}$, $\lambda_{4}$, $\lambda_{5} = [1.0, 1.0, 0.1, 0.1, 0.1]$ are weighting coefficients balancing the contribution of each loss term.

\section{Real-World Visual Comparison}  
\begin{figure*}[t]
    \centering
    \includegraphics[width=1\linewidth]{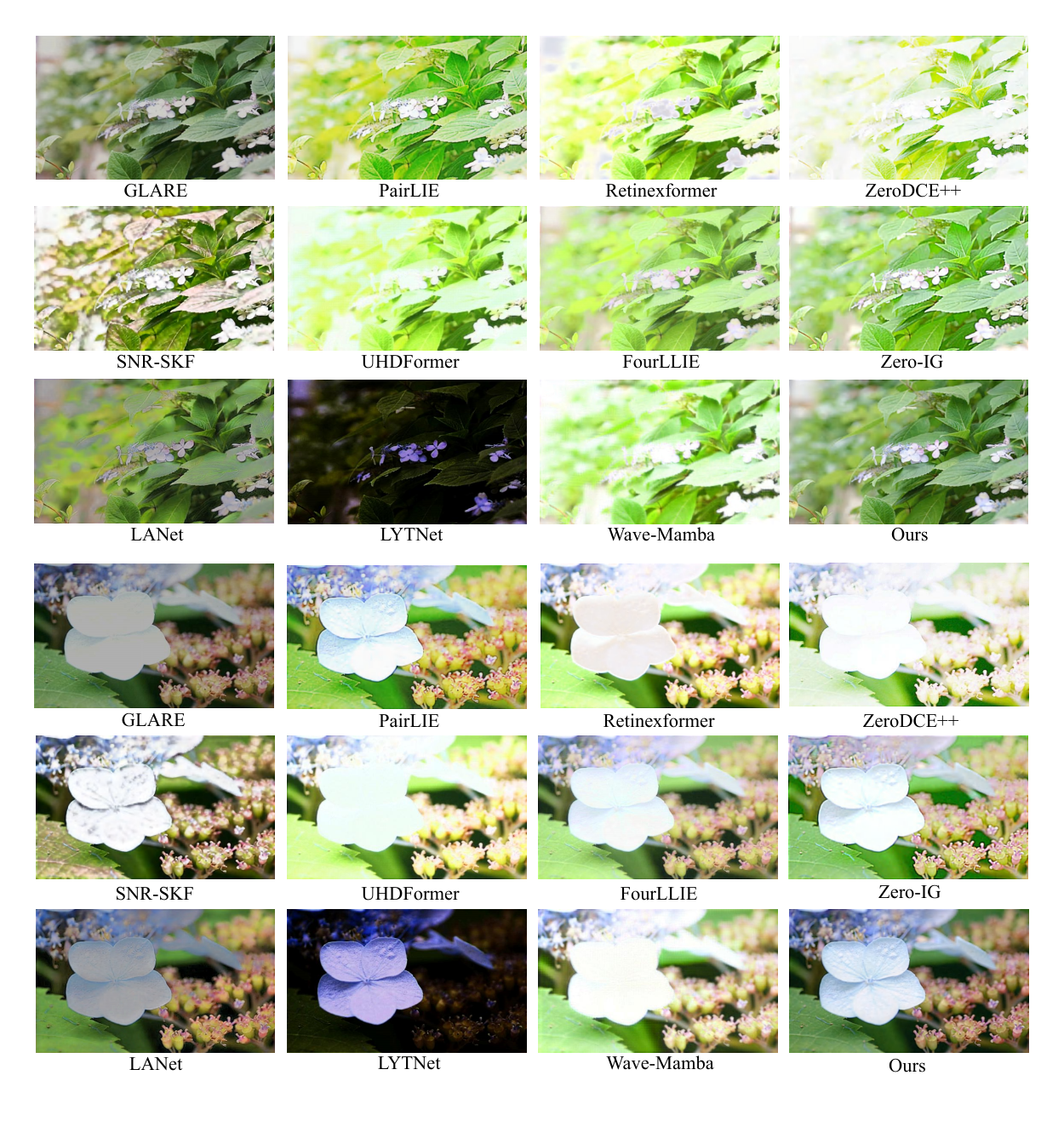}
    \caption{Visual comparison on DICM dataset.}
    \label{fig:com_DICM2}
\end{figure*}

\begin{figure*}[t]
    \centering
    \includegraphics[width=1\linewidth]{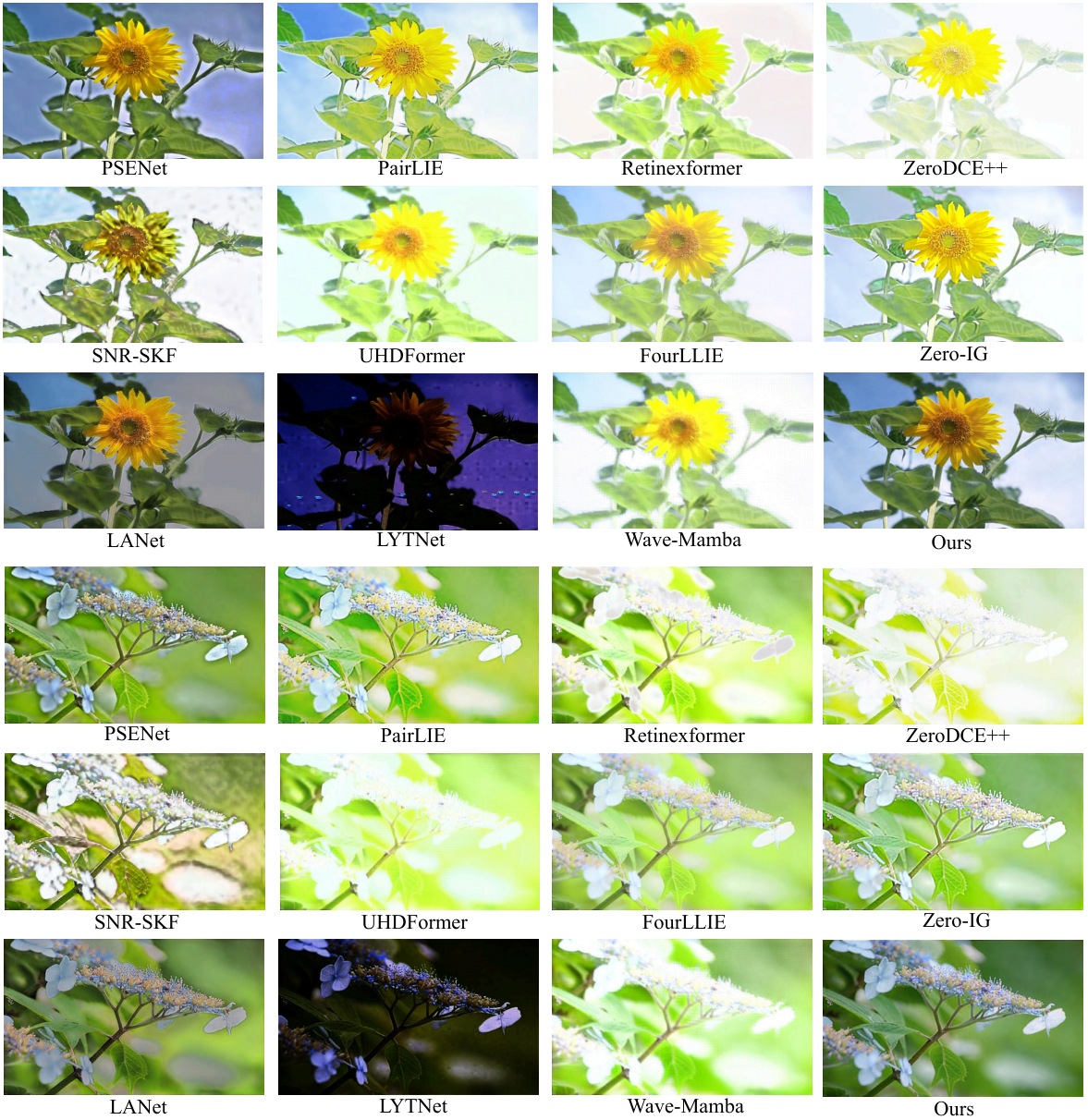}
    \caption{Visual comparison on DICM dataset.}
    \label{fig:com_DICM3}
\end{figure*}

\begin{figure*}[t]
    \centering
    \includegraphics[width=1\linewidth]{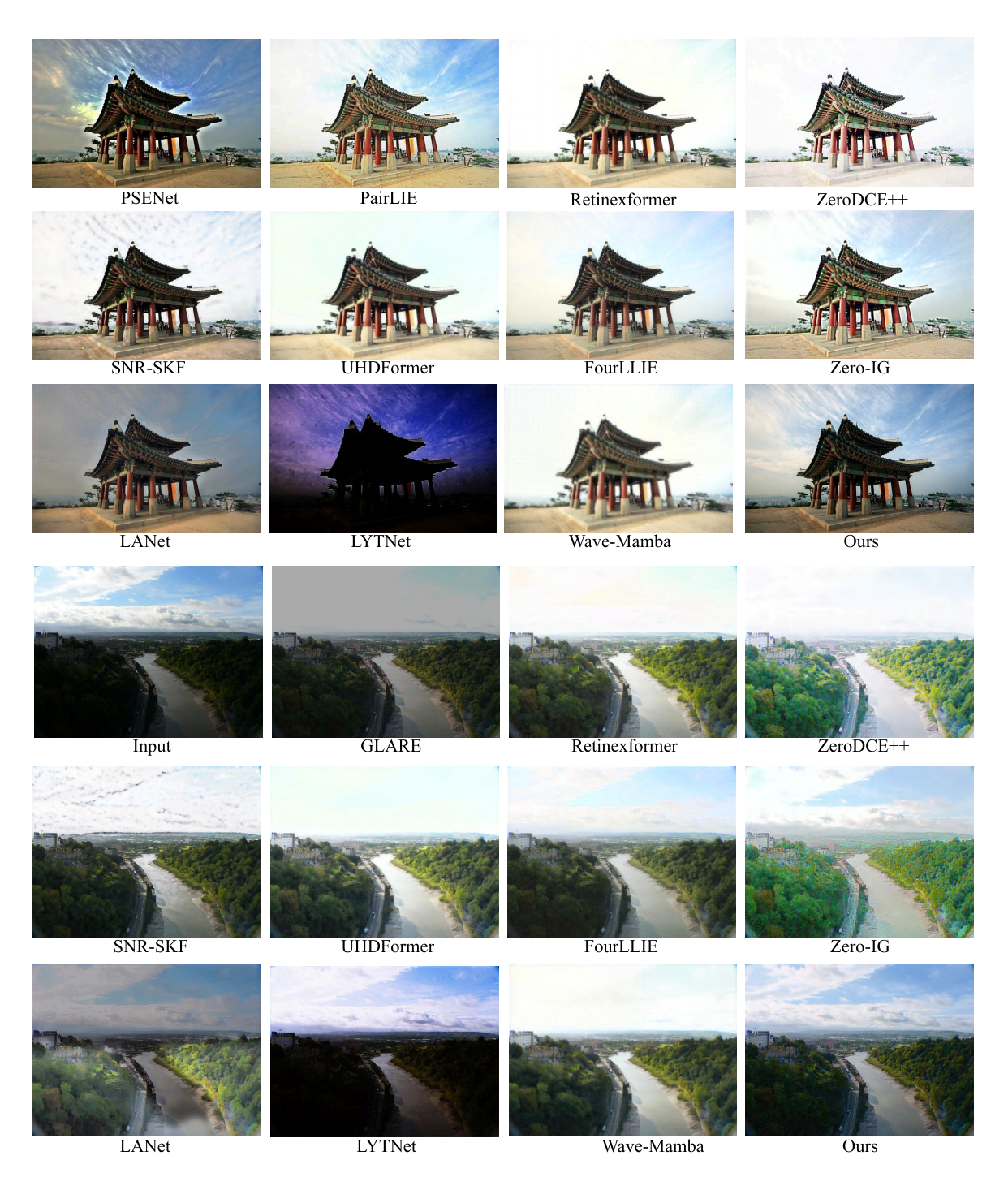}
    \caption{Visual comparison on DICM and VV datasets.}
    \label{fig:com_DICM1}
\end{figure*}

\begin{figure*}[t]
    \centering
    \includegraphics[width=1\linewidth]{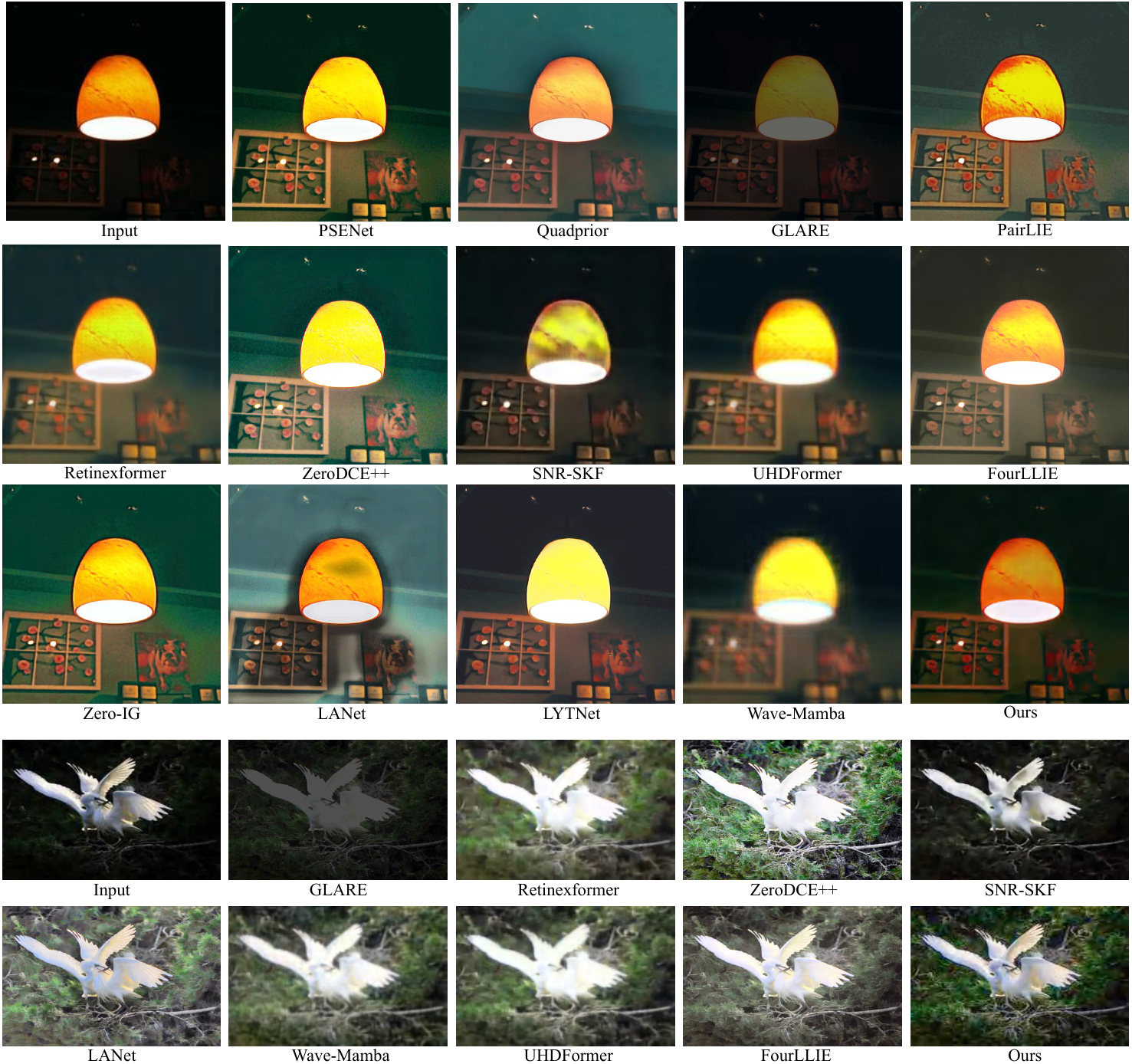}
    \caption{Visual comparison on LIME and NPE dataset.}
    \label{fig:com_vv}
\end{figure*}

\begin{figure*}[t]
    \centering
    \includegraphics[width=1\linewidth]{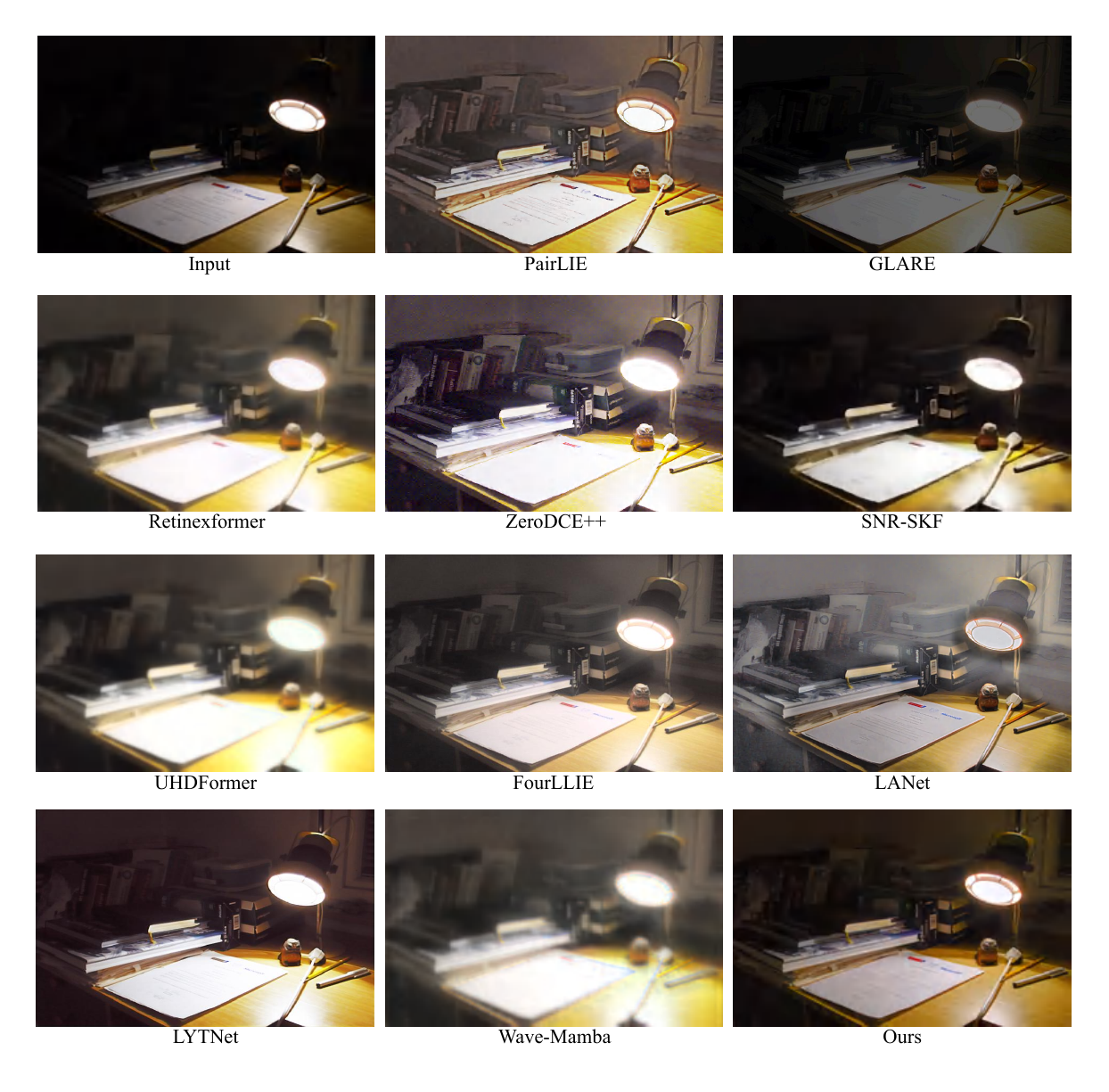}
    \caption{Visual comparison on MEF dataset.}
    \label{fig:com_npe}
\end{figure*}

As shown in Tab.~\ref{tab:niqe}, we conducted quantitative comparisons using the NIQE~\cite{6353522} metric on unpaired real-world datasets, where our method achieves state-of-the-art performance. To fully demonstrate our generalization capability on real-world scenarios, Fig.\ref {fig:com_DICM2}, Fig.\ref {fig:com_DICM3}, Fig.\ref {fig:com_DICM1}, Fig.\ref {fig:com_vv} and Fig.\ref {fig:com_npe} present visual comparisons with current state-of-the-art supervised and unsupervised methods on the DICM~\cite{dicm}, LIME~\cite{lime}, VV, MEF~\cite{mefl}, and NPE~\cite{npe} datasets. Compared methods include GLARE~\cite{zhou2024glare}, LYTNet~\cite{brateanu2401lyt}, LANet~\cite{yang2023learning}, FourLLIE~\cite{wang2023fourllie}, Wave-Mamba~\cite{zou2024wave}, Retinexformer~\cite{cai2023retinexformer}, UHDFormer~\cite{wang2024uhdformer}, SNR-SKF~\cite{wu2023learning}, ZeroDCE++~\cite{li2021learning}, PairLIE~\cite{fu2023learning}, PSENet~\cite{nguyen2023psenet} and Zero-IG~\cite{shi2024zero}. Our approach consistently achieves superior brightness and color recovery, resulting in optimal visual quality and image detail preservation across diverse challenging scenarios.

\end{document}